\documentclass[10pt]{article}
\PassOptionsToPackage{dvipsnames,svgnames,x11names}{xcolor}
\usepackage[citestyle=authoryear]{colab}
\usepackage{xspace}
\usepackage{marvosym}
\newcommand{\name}{EVA-Client\xspace}
\usepackage{footmisc}
\usepackage{calc}

\title{\protect\raisebox{-0.2em}{\protect\includegraphics[height=1.1em]{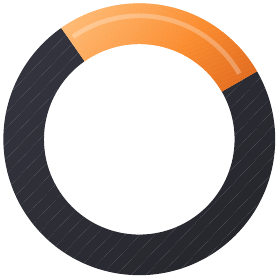}}\hspace{0.35em}EVA-Client: A Unified Framework for Deployment, Evaluation, and Data Collection on Real Robots}

\author{CoLab, Beihang University}

\labname{Colab}
\institution{}
\colabdate{July 2026}
\paperurl{https://colalab.net/projects/eva-client}
\githuburl{https://github.com/Noietch/EVA-CLIENT}
\colabrunningtitle{EVA-Client: A Unified Framework for Deployment, Evaluation, and Data Collection on Real Robots}

\begin{document}

\maketitle

\begin{colababstract}
Robot manipulation policies powered by Vision-Language-Action Models (VLAs), Video-Action Models (VAMs), and World-Action Models (WAMs) have made rapid progress with increasingly mature training frameworks and large-scale multimodal foundation models. Deploying a trained policy onto a physical robot, however, remains an underappreciated challenge in the community. Real-robot deployment often involves several tightly coupled aspects, including robot integration, real-time execution, and physical evaluation, making these policies difficult to reproduce, debug, and compare on physical robots.
We present \textbf{EVA-Client}, an open-source framework for deployment, debugging, data collection, and evaluation on real robots. It is a single client for the real-robot side of the iteration loop, from data collection and deployment through smoothing and evaluation, with the resulting records feeding external training. EVA-Client addresses the deployment problem in three aspects. First, it provides a component-decoupled, signal-source- and robot-agnostic architecture: supported robot backends, inference strategies, and transport middlewares form an orthogonal grid that runs out of the box, so adding a robot description or a strategy touches only the corresponding layer. Second, it makes deployment inspectable through Debug, Collect, and Evaluation workflows, with execution modes ranging from open-loop simulation and single-chunk stepping to continuous control. Third, it treats every evaluation as a data collection: each run records rollout data in a training-ready format together with comprehensive per-run logs and an interactive side-by-side viewer, so a reproducible evaluation can feed the next round of external training. EVA-Client further unifies major real-time inference strategies, including synchronous/asynchronous execution, ACT-style temporal ensembling, and Real-Time Chunking, together with a naive-async baseline for ablation, behind a single configuration surface, and is released as open infrastructure.
This report tracks the ongoing development of EVA-Client, with regular updates to capture new features and enhancements.

\vspace{0.4cm}
\colablinks
\end{colababstract}

\section{Introduction}

Embodied intelligence aims to build agents that can understand goals, perceive the physical world, and act through physical embodiments. Among its application domains, robot manipulation is a central testbed, because it requires grounding language and visual observations into precise, temporally coherent actions under contact-rich dynamics. Recent progress has therefore shifted from task-specific controllers toward generalist manipulation policies built on multimodal foundation models. Representative directions include Vision-Language-Action (VLA) models~\citep{rt2,openvla,pi0}, Video-Action Models (VAMs)~\citep{mimicvideo,UVAM}, and World-Action Models (WAMs)~\citep{genie-envisioner,lingbot-va,dreamzero,dreamdojo,GE-Sim-2}. These directions differ in whether actions are generated directly, through video-based representations, or through learned world dynamics. Around these models, the training ecosystem has become increasingly mature. Frameworks such as openpi~\citep{openpi}, LeRobot~\citep{lerobot}, VLA Foundry~\citep{vlafoundry}, and StarVLA~\citep{ye2026starvla,community2026starvla} provide modular model abstractions, shared data formats, benchmark integration, and scalable training recipes. These efforts have substantially lowered the barrier for training and reproducing robot policies.

\begin{figure*}[t]
\centering
\includegraphics[width=\textwidth]{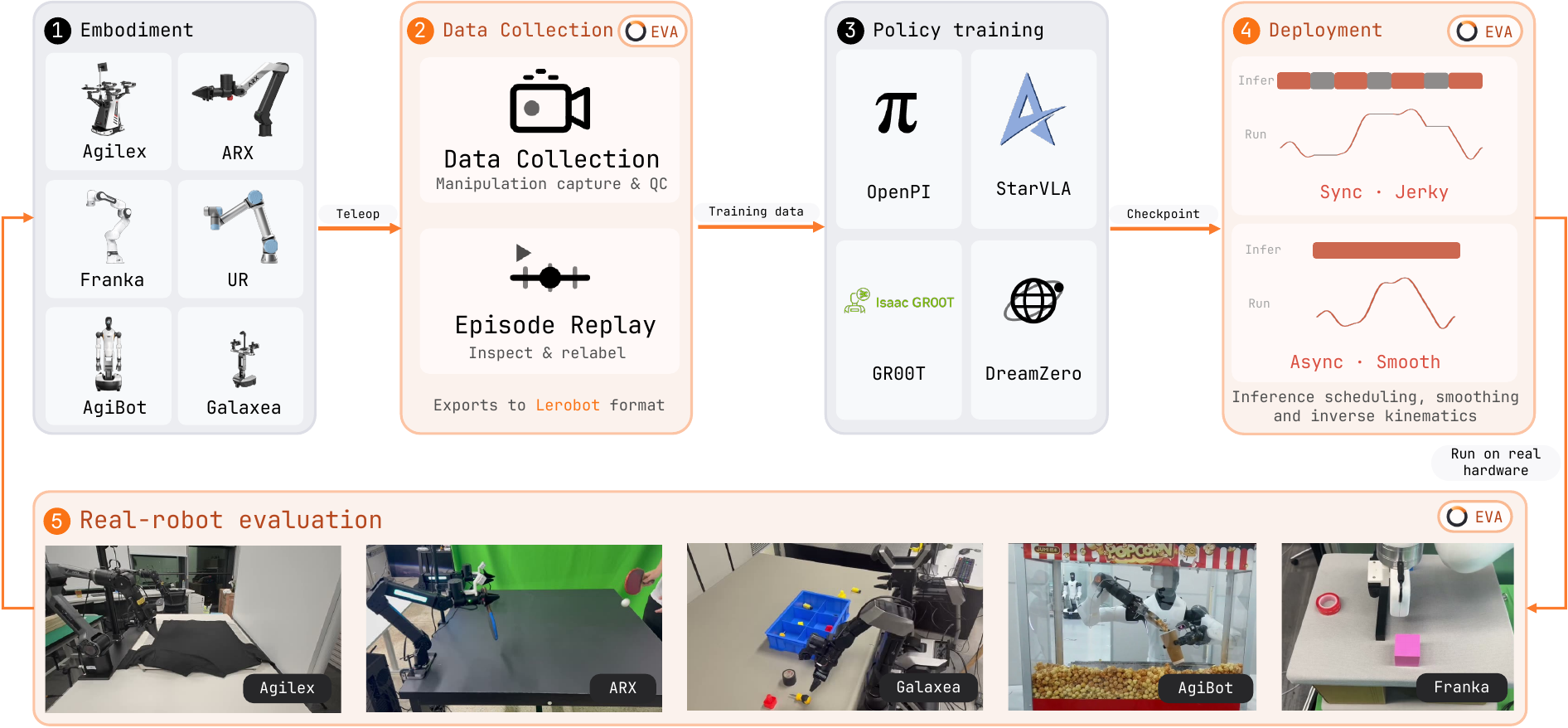}
\caption{\textbf{\name across the manipulation-policy lifecycle.} A left-to-right view of the broader pipeline around \name, spanning connected robot embodiments, data collection, external policy training, deployment, and real-robot evaluation. \emph{(1) Embodiment:} physical serial-arm platforms outside \name, namely \textbf{AgileX Piper, ARX R5, Franka, UR5e, AgiBot G2, and Galaxea R1-lite}, connect to one robot-agnostic backend through a single description class each (Section~\ref{sec:backends}). \emph{(2) Data collection:} a human teleoperates the robot while \name captures each demonstration, quality-checks it, and replays it for inspection, then exports it in the LeRobot format as training-ready data (Section~\ref{sec:collect}). \emph{(3) Policy training:} external frameworks such as openpi~\citep{openpi}, StarVLA~\citep{ye2026starvla,community2026starvla}, GR00T~\citep{gr00tn1_2025}, and Dream-Zero~\citep{dreamzero} consume the collected data and return a checkpoint; \name trains no models itself. \emph{(4) Deployment:} \name serves the trained checkpoint on real hardware, using inference scheduling, overlap smoothing, and inverse kinematics to turn action chunks into a smooth command stream rather than the jerky pause-and-go of naive synchronous execution (Section~\ref{sec:strategies}). \emph{Bottom band, real-robot evaluation:} the deployed policy is scored on physical robots under a standardized, fully logged protocol (Section~\ref{sec:eval}). \name covers the client-side data collection, deployment, and evaluation workflows; robot embodiments and policy training remain external.}
\label{fig:workflow}
\end{figure*}

Deployment, by contrast, is still largely underappreciated. Taking a checkpoint from ``it trained'' to ``it moves the robot correctly'' requires a deployment stack.
Beyond simply querying a policy server, this stack also needs to handle observation-command interfaces, real-time action scheduling, latency compensation, action-space conversion, and rollout logging.
Although these components are not usually treated as model contributions, they often determine whether a trained policy can run safely, smoothly, and reproducibly on physical hardware. In practice, they are still frequently implemented as method- or robot-specific scripts, for example the per-model deployment clients and evaluation entry points bundled with individual training stacks such as openpi~\citep{openpi} and LeRobot~\citep{lerobot}. This makes real-robot deployment difficult to inspect, reuse, and compare across policies and platforms.

This gap has three consequences. \textbf{Robot-integration} lies below the abstraction level of most model and training frameworks. A policy may share the same checkpoint format or inference API, but the control loop still depends on robot-specific cameras, state feedback, middleware, and action spaces. As a result, code built for one robot rarely transfers directly to another. \textbf{Real-time-execution} is coupled to deployment rather than training. Action chunking, asynchronous inference, latency compensation, and smoothing are often introduced as method-specific implementation details, so their behavior is hard to inspect or compare once moved onto physical hardware. \textbf{Physical-evaluation}, and, downstream of it, feeding what a deployment produces back into external training, is shaped by the deployment pipeline itself. Unlike training metrics or simulation scores, real-robot results depend on what the policy observed, what it predicted, how those predictions were transformed, and what commands were finally executed. This makes it hard to determine whether a policy improvement reflects a better model or a different deployment setup. Moreover, the demonstrations a physical run could yield are usually lost rather than routed back to the next round of external training.

To address these problems, we present \textbf{\name{}}, a unified client for inference, debugging, data collection, and deployment of trained manipulation policies. As Figure~\ref{fig:workflow} illustrates, \name{} sits within the broader policy lifecycle, spanning robot hardware integration, data collection, external policy training, real-time deployment, and real-robot evaluation, and unifies the real-robot stages of this cycle behind a single codebase. Positioned between signal sources and robot execution, it exchanges instructions, observations, and actions with sources such as openpi, StarVLA, GR00T, Dream-Zero, or human teleoperation, then turns the resulting actions into robot commands (Figure~\ref{fig:architecture}). The client commands the connected robot execution stack while exposing every stage of the deployment process to inspection, intervention, and evaluation. The same framework accommodates tasks with sharply different demands, from highly dynamic control to long-horizon execution. Across this range, even a single change in inference strategy can determine whether a task succeeds or stalls (Section~\ref{sec:strategies}, Figure~\ref{fig:teaser}). We make three contributions, which together close the deployment gaps identified above:

\begin{itemize}
    \item \textbf{A single client that covers the entire real-robot iteration loop.} Where the training side has converged onto shared frameworks such as openpi~\citep{openpi}, LeRobot~\citep{lerobot}, StarVLA~\citep{ye2026starvla,community2026starvla}, and VLA Foundry~\citep{vlafoundry}, the real-robot side is still a scatter of per-policy, per-robot scripts. \name closes this last gap by bringing data collection, deployment, chunk smoothing, evaluation, and the recorded feedback into one codebase. A checkpoint can therefore reach closed-loop iteration on real hardware without rewriting the deployment stack for each policy or robot.

    \item \textbf{A component-decoupled deployment architecture that works out of the box.} Signal sources, transport middlewares, robot descriptions, and inference strategies sit behind narrow interfaces, so any policy server or teleoperation stream can compose with any supported robot over any supported transport. Adding a robot, a strategy, or a transport touches only its own layer; bringing up a new deployment is a configuration choice, not a coding task.

    \item \textbf{Reproducible evaluation with a closed loop back to training.} Every evaluation run doubles as a data collection: EVA-Client records full rollout data in training-ready format during each run, alongside exhaustive per-run logs and a read-only comparison viewer. Physical runs therefore remain auditable and reproducible, and each evaluation automatically produces the material for the next training iteration rather than ending as an unrecorded impression.
\end{itemize}

The remainder of this report elaborates these contributions section by section. Section~\ref{sec:overview} presents the layered architecture, Section~\ref{sec:backends} the transport and robot hardware layer with its inverse kinematics, and Section~\ref{sec:modes} the operation model and its execution modes. Section~\ref{sec:strategies} covers the inference strategies, Section~\ref{sec:collect} the data-collection mode, and Section~\ref{sec:eval} the evaluation and logging subsystem, before the limitations and roadmap.

To be clear about scope, EVA-Client is a client and deployment layer for trained manipulation policies. It assumes a policy server already exists and is model-agnostic with respect to that server; it is not itself a policy, a benchmark, or a dataset, and its inverse-kinematics support targets serial-arm manipulators. We position EVA-Client as the deployment counterpart to the now-mature training frameworks and as complementary infrastructure that the field needs but has not consolidated.

\section{Related Work}
\label{sec:related}

EVA-Client sits at the intersection of three lines of work: the policy models it serves, the inference techniques it consolidates, and the embodied infrastructure it builds on.

\paragraph{Policy techniques.} Manipulation-policy models have evolved rapidly along several lines. Early language-conditioned robotic transformers cast control as sequence prediction over discretized actions~\citep{rt1,rt2}, and subsequent open Vision-Language-Action models such as OpenVLA released weights and recipes that made this family broadly reproducible~\citep{openvla}. A parallel line replaced autoregressive decoding with generative action heads that use diffusion or flow matching to emit smooth action sequences~\citep{diffusionpolicy,pi0}. More recently, Video-Action Models predict actions through learned video representations~\citep{mimicvideo,UVAM}, while World-Action Models couple a learned world model with action generation so that rollouts can be planned or imagined before execution~\citep{genie-envisioner,dreamzero,GE-Sim-2}. These families differ in how actions are produced, yet each ultimately returns a horizon of future actions that a deployment layer must schedule and execute on real hardware.

\paragraph{Inference techniques.} Because chunk-based policies return a horizon of future actions rather than a single command, real-time execution must decide how to schedule successive queries and reconcile overlapping predictions. Action chunking with temporal ensembling was popularized by ACT~\citep{act}, which blends overlapping chunks through an exponentially weighted average, and diffusion policies likewise emit short action sequences per query~\citep{diffusionpolicy}. To keep the robot moving during inference, asynchronous prefetch runs the policy in a background thread and blends each arriving chunk over its overlap window~\citep{sima2026kai0}, while Real-Time Chunking conditions generation on the previously committed actions so that successive chunks agree at the source~\citep{rtc}. Each technique is individually well motivated, but they are usually implemented in isolation and tied to one model, which makes them hard to compare. EVA-Client re-implements them behind a single configuration surface, so scheduling and smoothing become a switchable variable that can be configured and compared on the same robot.

\paragraph{Embodied infrastructure.} EVA-Client is built on the surrounding embodied-systems ecosystem rather than replacing it. Robot middleware such as ROS~\citep{ros} and ROS2~\citep{ros2} moves observations and commands, kinematics is handled by PyRoki~\citep{pyroki}, and the interactive console renders through Viser~\citep{viser}. In this way, transport, inverse kinematics, and visualization reuse mature community components. The teleoperation and hardware lineage of these components traces back to low-cost bimanual platforms such as ALOHA~\citep{aloha}. Training and data stacks such as LeRobot~\citep{lerobot}, openpi~\citep{openpi}, StarVLA~\citep{ye2026starvla,community2026starvla}, and VLA Foundry~\citep{vlafoundry} consolidate model abstractions, shared data formats, and scalable recipes, but ship per-model deployment and evaluation scripts as adjuncts to training. EVA-Client is deliberately orthogonal: it provides the deployment, debugging, and physical-evaluation layer these stacks leave as per-model glue, and reads the LeRobot dataset format~\citep{lerobot} for offline replay so the two compose rather than overlap.

\section{Framework Overview}
\label{sec:overview}

\begin{table*}[t]
\centering
\caption{Debugging and execution modes. Modes progress from safe and fully observable open-loop simulation to fully real-time execution on hardware. Evaluation (Section~\ref{sec:eval}) runs on top of these modes; data collection (Section~\ref{sec:collect}) reuses the same console and backends.}
\label{tab:modes}
\begin{tabular}{@{}lllm{6.2cm}@{}}
\toprule
\textbf{Mode} & \textbf{Drives} & \textbf{Granularity} & \textbf{Primary use} \\
\midrule
Open-loop simulation     & Sim      & Continuous   & Inspect policy behavior with no hardware risk \\
Real single-chunk stepping   & Real robot       & One chunk    & Step through execution on hardware, chunk by chunk \\
Segmented sim-to-real    & Sim then real    & One chunk    & Preview a chunk in sim, then confirm it on the real robot \\
Continuous execution     & Real robot       & Continuous   & Real-time deployment at full control rate \\
Data collection          & Real robot       & Continuous   & Record teleoperated demonstrations into training-ready datasets \\
\bottomrule
\end{tabular}
\end{table*}

\begin{figure*}[t]
\centering
\includegraphics[width=\textwidth]{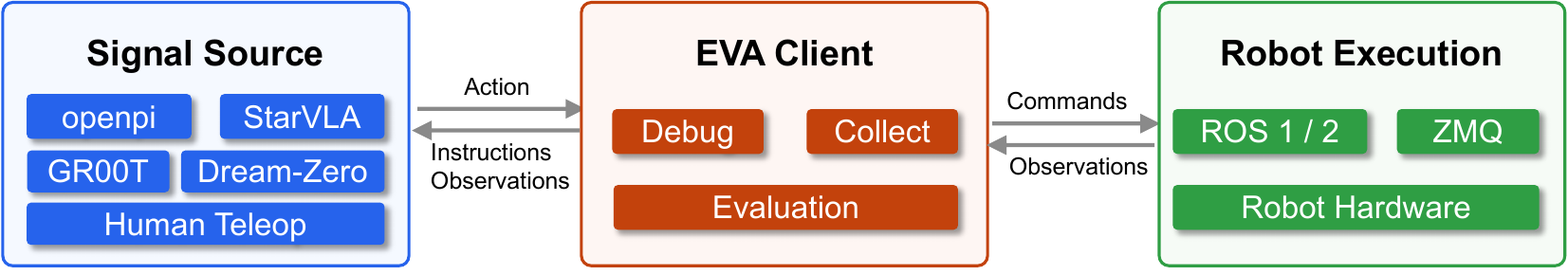}
\caption{\textbf{Where \name sits in the deployment pipeline.} \name mediates between signal sources and robot execution. \emph{Left:} signal sources include trained policy systems such as openpi~\citep{openpi}, StarVLA~\citep{ye2026starvla,community2026starvla}, GR00T~\citep{gr00tn1_2025}, and Dream-Zero~\citep{dreamzero}, as well as human teleoperation. Model-based sources receive instructions and observations from \name and return actions or action chunks; teleoperation provides actions directly. \emph{Center:} \name exposes this process through Debug, Collect, and Evaluation workflows. \emph{Right:} robot execution backends turn actions into robot commands and return synchronized observations over ROS~1/2 or ZMQ before reaching the robot hardware (Section~\ref{sec:backends}). The source and execution sides remain narrow and replaceable, so a new policy source, teleoperation source, or robot backend can be added without rewriting the whole deployment stack.}
\label{fig:architecture}
\end{figure*}

\name is built around three design principles, which directly target the three deployment gaps identified in the Introduction. At a high level (Figure~\ref{fig:architecture}), \name is a thin client between signal sources and robot execution. Signal sources include trained policy servers such as openpi, StarVLA, GR00T, and Dream-Zero, as well as human teleoperation. The client sends instructions and observations to model-based sources, receives actions or action chunks in return, and also accepts teleoperated actions for collection. It then issues commands to the robot execution stack over ROS~1/2 or ZMQ and reads back synchronized observations. The deployment process is exposed through Debug, Collect, and Evaluation workflows, with additional Replay and Result views detailed in Sections~\ref{sec:modes} to~\ref{sec:eval}. 

The four execution granularities of Table~\ref{tab:modes} live inside the Debug workflow. (1) Robot-agnostic: the control loop, inference strategies, and debugging modes do not assume a particular robot, middleware, or action representation. (2) Reproducible: a deployment is fully described by a configuration, and every run can record exactly what was observed, inferred, and executed. (3) Observable: the user can inspect and intervene at each stage in the client, preview in simulation, step one chunk at a time, and visualize trajectories in the real world, without rewriting code.

Internally, the client is organized into five layers with narrow interfaces. Every workflow drives these same layers, differing only in how the CLI/web front-end presents them. The transport layer abstracts how synchronized observations are captured and how commands are delivered, whether from a real robot, a dataset, or a socket-based fake node. The robot description layer declares actuator groups, cameras, the observation schema, topic mappings, and optional kinematics. The policy client queries a model-based signal source and returns an action chunk, while teleoperation sources provide actions through the collection path. The inference strategy decides when to call the policy and how to turn overlapping chunks into a single smooth action stream. The CLI/web layer drives the session state machine and provides the interactive, collection, and evaluation front-ends. Because the layers communicate through small dataclass interfaces, namely a shared observation interface, action arrays or chunks, and a robot-description object, each can be replaced independently.

\paragraph{Control loop.} The control loop is the closed cycle of Figure~\ref{fig:architecture}. On each iteration the client reads a synchronized observation back from the robot execution stack. For model-based sources, it forwards that observation together with the language instruction to the policy server, receives an action chunk, and issues the resulting commands to the robot. For teleoperation, the source supplies operator actions that go through the same execution and logging path. The robot then produces the next observation and closes the loop. This cycle runs at a fixed control rate that is deliberately decoupled from the rate at which a policy is queried. How successive chunks are scheduled and blended into the command stream is the concern of the inference strategy (Section~\ref{sec:strategies}). A single session state machine, cycling from idle to ready to running, governs setup, start, stop, and reset uniformly across every workflow.

\paragraph{Configuration.} A deployment is fully specified declaratively: one configuration file fixes the robot, transport backend, policy endpoint, action-space modes, loop rates, prompts, and inference strategy. Command-line flags override individual fields so the same configuration can be retargeted without edits. This makes a run reproducible from its configuration alone and is what the evaluation subsystem (Section~\ref{sec:eval}) records to reconstruct it later.

\section{Backends and Hardware}
\label{sec:backends}

The architecture of \name keeps the control loop robot-agnostic, supporting new hardware or a new observation source reduces to swapping a single layer. This section describes the transport backends that move observations and commands, the robot descriptions that declare a platform, and the action-space and inverse-kinematics machinery that lets policies and robots disagree about representation.

\paragraph{Transport backends.} EVA-Client decouples the control logic from how observations and commands actually move.

The ROS1 backend subscribes to camera and joint-state topics and time-synchronizes them into observation frames. Rather than relying on the ROS message-synchronization utility with a tolerance window, it buffers each topic in a bounded deque whose length is a configurable staleness window. On each request, it aligns to the latest timestamp common to all required streams, dropping every message older than that instant. If no timestamp is common to all required streams within the current buffers, the request waits until the streams re-synchronize rather than emitting a mismatched frame. It then publishes joint commands to the real robot and, optionally, to parallel simulation topics.

The ROS2 backend mirrors this design on the ROS2 client library through per-topic buffering, header-stamp alignment, and best-effort QoS for live streams. This lets ROS2-native robots be handled directly rather than bridged.

The dataset backend replays a LeRobot-format episode~\citep{lerobot}, comprising parquet observations plus MP4 video, so that an inference pipeline can be exercised open-loop with no hardware. The ZMQ backend speaks a lightweight wire protocol over a socket; pointed at a bundled fake execution-layer node, it synthesizes observation frames in a separate process, enabling development without a physical robot. Removing the dependency on a policy server is handled separately, by a bundled fake policy server or by the server-free mock/replay policy clients, not by the transport layer. ROS is required only for the matching ROS backend; the ZMQ and dataset backends have no ROS dependency.

\paragraph{Robot support.} A robot is declared, not coded: a robot-description object lists actuator groups such as arms, grippers, and a base, together with the camera/observation schema, topic mappings, and optional kinematics, and registers itself in a registry. The reference implementation is a dual-arm Piper, namely two 6-DoF arms plus grippers, for a 14-D action vector $=2\times(6\text{ joints}+1\text{ gripper})$. The same generic backend already connects to a heterogeneous supported hardware set: the \textbf{Franka}, \textbf{UR5e}, \textbf{Galaxea R1-lite}, \textbf{AgileX Piper}, \textbf{AgiBot G2}, and \textbf{ARX R5} (with more platforms on the roadmap). Inverse kinematics is provided by a shared, robot-agnostic PyRoki~\citep{pyroki} solver, detailed below. Because every other subsystem talks to the robot only through this interface, supporting a new robot means only writing one description class.

\paragraph{Camera support.} For ROS-based robots, cameras are ordinary topics consumed by the transport layer. Robots that are not ROS-based need an alternative image source; a middleware-independent camera interface for non-ROS robots is a current limitation (Section~\ref{sec:roadmap}).

\paragraph{Action spaces and inverse kinematics.} EVA-Client decouples three action spaces that are usually conflated: the observation state space, the policy output space, and the publication space. Each may independently be joint or end-effector. In end-effector mode each arm is represented as a position, an orientation, and a gripper command. The orientation is accepted as a quaternion, roll-pitch-yaw, or 6D rotation and canonicalized internally. The scalar gripper dimension can optionally be snapped to open/closed at a configurable threshold for grippers driven as binary open/close, while continuous grippers pass their command through unchanged. When the policy outputs end-effector poses but the robot is commanded in joint space, the framework invokes the IK solver to convert poses to joint targets on the fly. This makes it possible, for example, to feed joint-space observations to a policy that predicts end-effector poses and still command a joint-controlled arm, without changing the policy or the robot driver.

\paragraph{Continuous IK.} The IK solver builds on PyRoki~\citep{pyroki}, a JAX-based kinematics toolkit, using its nonlinear least-squares solver. Both inverse and forward kinematics come from PyRoki's differentiable forward-kinematics routine, and one backend serves every robot in the zoo.

Each frame of an action chunk is a single-configuration least-squares problem solved by Levenberg--Marquardt, using a trust-region step with dense-Cholesky factorization. It minimizes three weighted costs under a hard joint-limit constraint: an end-effector pose cost with analytic Jacobian and separate position/orientation weights, a rest cost biasing toward the home pose, and a velocity cost penalizing only over-speed steps.

Continuity is maintained by solving frames sequentially. Each frame is warm-started from the previous solution, and the first frame is anchored to the measured joint state so execution begins without a jump. With the velocity cost, this biases each solve toward the nearby IK branch and keeps targets on a mostly continuous joint trajectory; near singularities or joint limits discontinuities can still occur and are logged.

Cost weights, namely separate position/orientation, rest, and velocity terms, and the tracking-error tolerance are exposed as configurable solver parameters; a frame whose tracking error exceeds the tolerance is still commanded but is flagged for review rather than rejected. This complements the inference-strategy smoothing of Section~\ref{sec:strategies}: continuous IK keeps joint targets on a coherent trajectory, while the inference strategies blend overlapping chunks into a smooth command stream.

\section{Operation Model}
\label{sec:modes}

\begin{figure*}[t]
\centering
\includegraphics[width=\textwidth]{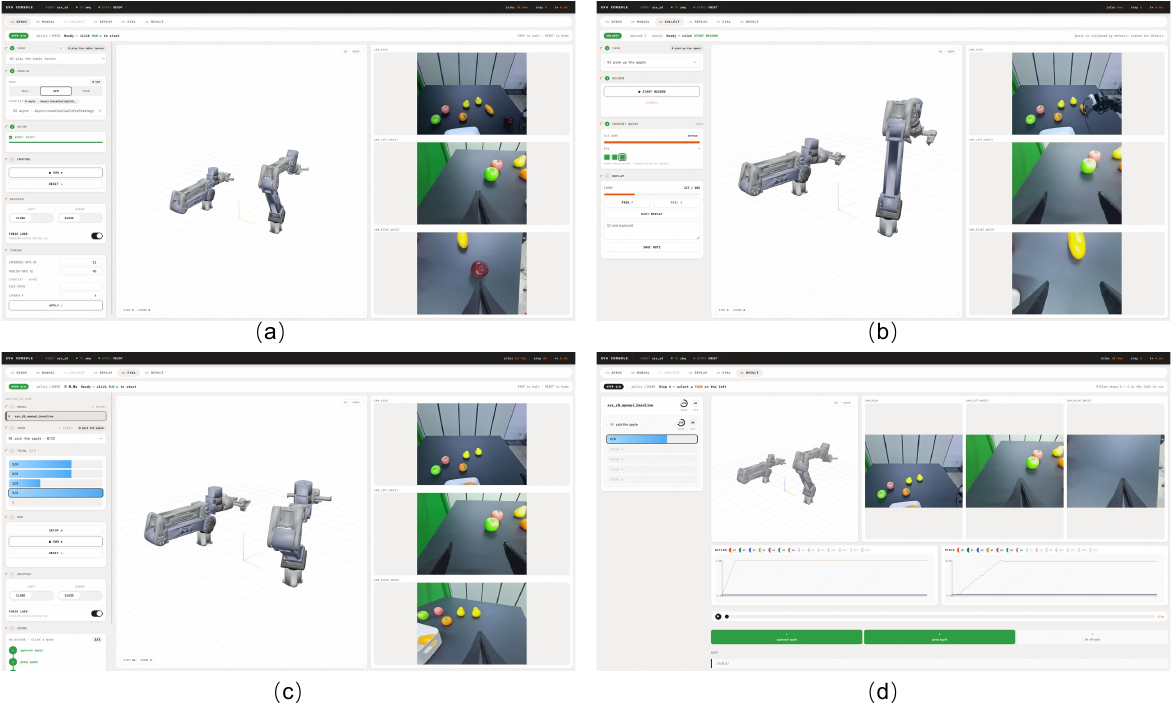}
\caption{\textbf{The \name web console across its usage modes.} A single browser interface drives the whole deployment. A persistent tab bar selects among five tabs, namely the three primary usage modes Debug, Collect, and Eval, plus a Replay tab and a read-only Result viewer (Section~\ref{sec:eval}); a status bar reports the active robot, transport, session state, and per-call inference latency. Every mode shares the same layout: a left control panel, a center Viser-based~\citep{viser} 3D scene rendering the live dual-arm state and executed trajectories, and, on the right, the time-synchronized camera streams, one overhead and two wrist views, that form each observation frame. This 3D scene also doubles as the rendering target for the open-loop and sim-to-real modes. \emph{(a) Debug:} selects the execution mode (\textsc{real}/\textsc{sim}/\textsc{step}) and inference strategy, then runs the policy, here in open-loop simulation, with control/inference rates and the latency cap $k_{\max}$, the maximum number of stale actions to discard (Section~\ref{sec:strategies}), exposed as tuning fields. \emph{(b) Collect:} records a teleoperated episode into a training-format dataset, then replays it for pass/fail QC with a free-text note (Section~\ref{sec:collect}). \emph{(c) Eval:} runs the scored-trial protocol of Section~\ref{sec:eval}, with a model label, a per-trial success tally, and a milestone checklist scored live. \emph{(d) Result:} the read-only result viewer, showing per-checkpoint statistics and side-by-side checkpoint comparison (Section~\ref{sec:eval}). The Replay tab reuses the same open-loop viewer to review recorded episodes: dataset replay (Section~\ref{sec:backends}) and Collect quality control (Section~\ref{sec:collect}). The same console thus spans inspection, data collection, and evaluation without leaving the page.}
\label{fig:console}
\end{figure*}

EVA-Client treats deployment as a debugging activity rather than a one-shot launch. The framework therefore exposes its whole operation model through a single web console (Figure~\ref{fig:console}). A persistent tab bar selects among five tabs, namely the three primary usage modes \textbf{Debug}, \textbf{Collect}, and \textbf{Eval}, plus a \textbf{Replay} tab and a read-only \textbf{Result} viewer. All of these tabs share one layout: a control panel, a Viser-based~\citep{viser} 3D scene, and the time-synchronized camera streams that form each observation frame. This section describes the inspection and execution modes; the Collect data-collection mode and the Eval and Result evaluation modes are detailed in Sections~\ref{sec:collect} and~\ref{sec:eval}. Within the Debug tab, the execution modes trade off safety and observability against realism, summarized in Table~\ref{tab:modes}. Throughout, the Viser-based 3D scene renders joint state, end-effector poses, and executed trajectories live, and doubles as the rendering target for the open-loop and sim-to-real modes below.

\paragraph{Open-loop simulation.} The policy runs against live or synthetic observations, but actions are routed exclusively to a Viser-based 3D visualization (the center view of Figure~\ref{fig:console}), rather than to physical hardware. This is the safest entry point: a user can confirm that a checkpoint produces sane motion before any physical risk is incurred.

\paragraph{Real single-chunk stepping.} On the physical robot, the user advances exactly one action chunk per command. Between chunks the robot is stationary and the user inspects the result, making it possible to localize failures, such as a bad grasp pose, to a specific inference step rather than watching a continuous run fail and guessing why.

\paragraph{Single-chunk sim-to-real stepping.} This mode interleaves the previous two: each chunk is first previewed in simulation; the user then either commits it to the real robot or cancels it. It provides a human checkpoint between prediction and physical execution, which is valuable when validating a new checkpoint or operating near fragile objects.

\paragraph{Continuous execution.} The full real-time deployment mode: the control loop publishes actions continuously at the configured rate, with the selected inference strategy (Section~\ref{sec:strategies}) handling chunk scheduling and smoothing in the background. This is the mode used for actual task execution.

\section{Inference Strategies}
\label{sec:strategies}

\begin{figure*}[t]
\centering
\includegraphics[width=\textwidth]{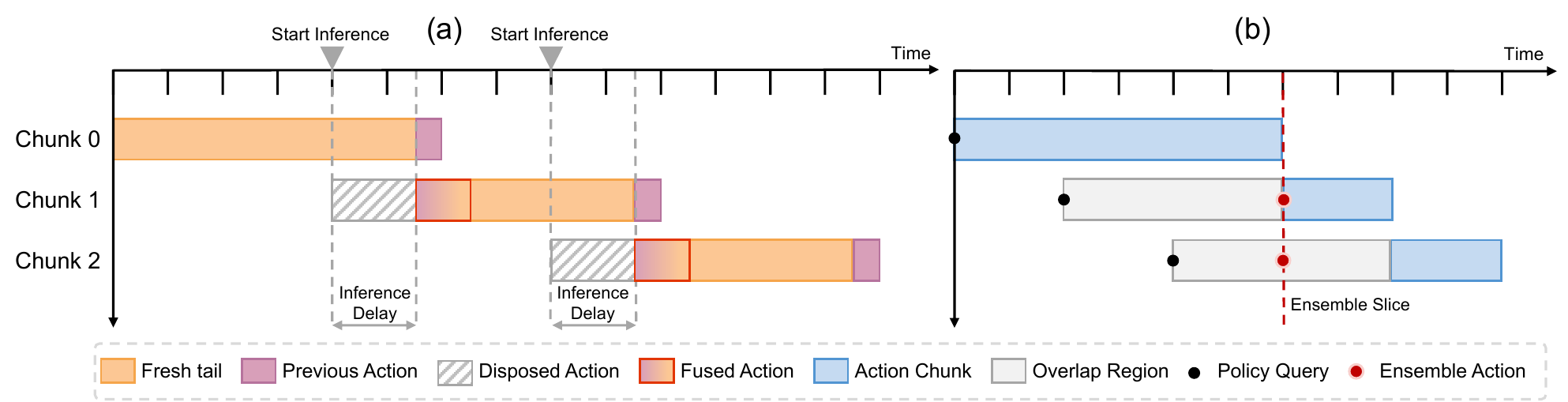}
\caption{\textbf{Two asynchronous smoothing strategies on a shared timeline.} Both panels read left-to-right in time, with successive inference calls stacked top-to-bottom; each call starts from an observation (marked \emph{Start Inference} in (a), \emph{Policy Query} in (b)) and returns only after an \emph{inference delay}, so consecutive chunks overlap in time. \emph{(a) Asynchronous prefetch with linear-overlap blending:} when a new chunk arrives, its leading actions (\emph{disposed}) are dropped to compensate for the steps already executed during the inference delay. The overlap window (\emph{fused}) is linearly blended between the \emph{Previous Action} and the new prediction, and the remaining \emph{fresh tail} is appended unchanged. \emph{(b) ACT-style temporal ensembling:} rather than discarding earlier predictions, every chunk that covers the current control timestep contributes to the executed action. At each \emph{ensemble slice} the overlapping predictions are combined by an exponentially weighted average over all chunks active at that instant into the executed \emph{Ensemble Action} (red). Following ACT, the weighting favors earlier (older) predictions; with the default decay the weights are nearly uniform.}
\label{fig:inference-strategy}
\end{figure*}

Real-time execution of chunk-based policies requires more than repeatedly querying the model. Since each inference call returns a horizon of future actions while incurring non-negligible latency, consecutive chunks overlap in time and may disagree at their boundaries. The inference strategy must therefore decide when to request new chunks, which delayed actions should still be trusted, and how overlapping predictions should be combined into a smooth command trajectory. \name implements representative strategies for this problem behind a unified interface; Table~\ref{tab:strategies} summarizes their differences, and Figure~\ref{fig:inference-strategy} contrasts the two overlap-handling schemes on a shared timeline. We denote the action chunk produced at inference step $j$ as $A^{(j)} = (a^{(j)}_0, a^{(j)}_1, \dots, a^{(j)}_{H-1})$, where $H$ is the prediction horizon and each $a \in \mathbb{R}^d$.

\begin{figure*}[t]
\centering
\begin{subfigure}{\textwidth}
\centering
\includegraphics[width=\textwidth]{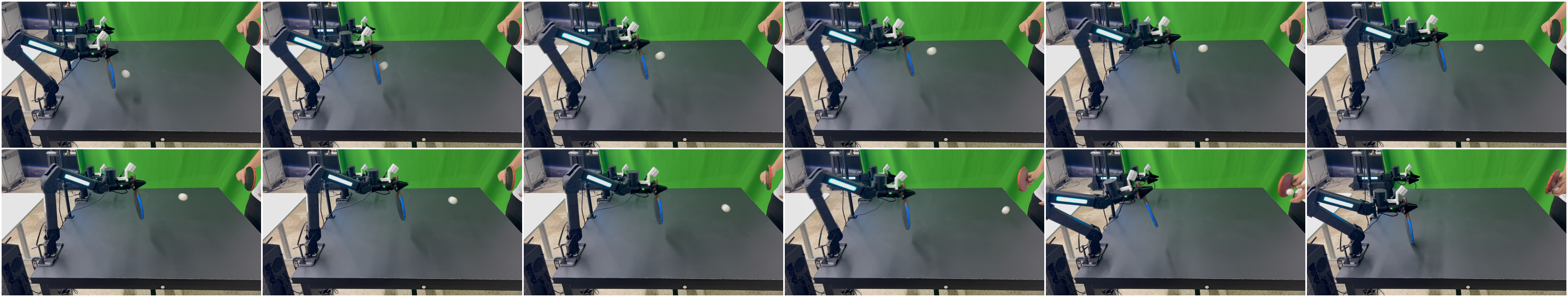}
\caption{\textbf{Table tennis, a high-dynamics task.} A Piper arm returns a ball, shown as a filmstrip whose frames advance left-to-right, top-to-bottom. The task stresses real-time responsiveness: under synchronous execution the pause-and-go stall keeps the arm from tracking the fast ball, so the rally does not get going in our deployment, whereas the rally shown here runs under asynchronous scheduling. \name keeps the control loop running while inference proceeds in a background thread, using latency-compensated chunk scheduling so that fast, reactive motions stay aligned with live observations.}
\label{fig:teaser-pingpong}
\end{subfigure}

\vspace{0.6em}

\begin{subfigure}{\textwidth}
\centering
\includegraphics[width=\textwidth]{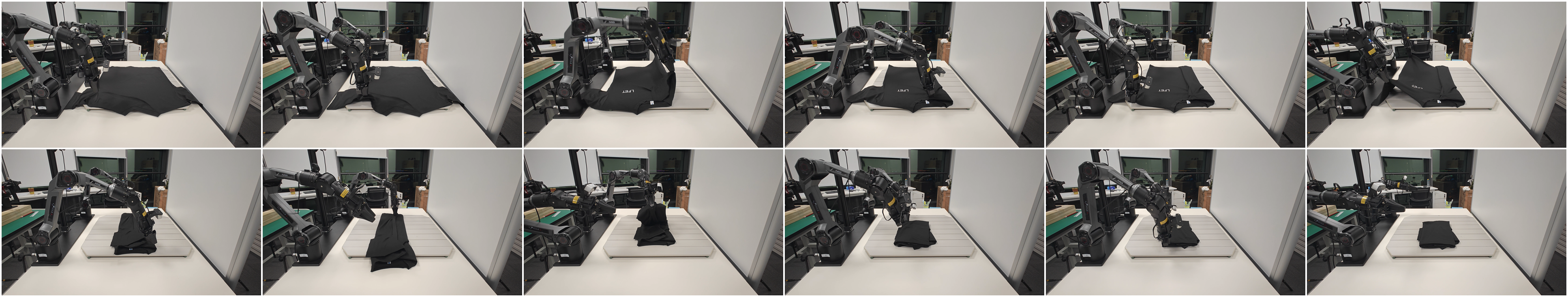}
\caption{\textbf{Cloth folding, a long-horizon stability task.} The same Piper arm folds a cloth on a tabletop, progressing spread $\rightarrow$ folded; the filmstrip frames advance left-to-right, top-to-bottom. The task stresses temporal coherence over many steps: \name blends overlapping action chunks into a smooth command stream, keeping the trajectory stable across the extended sequence. This fold runs under asynchronous execution, whose overlap blending keeps the long-horizon trajectory stable.}
\label{fig:teaser-fold}
\end{subfigure}
\caption{\textbf{One robot-agnostic framework, two contrasting demands.} Each panel is a time-ordered filmstrip read left-to-right, top-to-bottom. \emph{Top:} table tennis demands high dynamics: fast, reactive control in real time. \emph{Bottom:} cloth folding demands long-horizon stability: a smooth, coherent trajectory over an extended sequence. The same \name deployment drives both, changing only its inference strategy and robot description rather than any task-specific code. For the high-dynamics rally, synchronous pause-and-go stalls the arm and the rally does not get going in our deployment, so an asynchronous strategy is required; the folding task likewise runs under asynchronous execution (Section~\ref{sec:strategies}).}
\label{fig:teaser}
\end{figure*}

Why this choice matters in practice is easiest to see on real tasks with opposite demands. Figure~\ref{fig:teaser} shows the same \name deployment driving two of them, high-dynamics table tennis (Figure~\ref{fig:teaser-pingpong}) and long-horizon cloth folding (Figure~\ref{fig:teaser-fold}), where only the inference strategy and robot description change, not any task-specific code. Under synchronous execution the arm pauses for every forward pass, and this pause-and-go stall keeps it from tracking the fast ball, so the rally does not get going in our deployment. Asynchronous scheduling keeps the control loop running during inference, which lets the rally proceed. The folding task runs under asynchronous execution as well. These are illustrative observations from real deployments rather than a controlled benchmark. The strategies formalized below are selectable per task from a single configuration surface.

\begin{table}[t]
\centering
\caption{Inference strategies implemented in EVA-Client. ``Async'' indicates inference overlaps execution in a background thread. Smoothing shorthands: linear overlap, latency-trimmed chunks blended over their overlap window; exp.\ weighted, ACT-style exponentially weighted temporal ensembling; chunk replace, newest chunk replaces the buffer with no blending; overlap+server, RTC server-side conditioning plus a final linear-overlap pass.}
\label{tab:strategies}
\begin{tabular}{@{}lcl@{}}
\toprule
\textbf{Strategy} & \textbf{Async} & \textbf{Smoothing} \\
\midrule
Synchronous          & no  & - \\
Async prefetch       & yes & Linear Overlap \\
Temporal Ensemble        & yes & Exponentially Weighted \\
Naive Asynchronous         & yes & Chunk Replace \\
Real-Time Chunking (RTC) & yes & Overlap + Server \\
\bottomrule
\end{tabular}
\end{table}

\paragraph{Synchronous execution.}
The client requests a chunk, optionally crops it to an execution horizon, executes all of its actions in order, then requests the next chunk. Control is simple and exactly reproducible, but the robot pauses during every forward pass, producing pause-and-go motion. This mode is most useful for debugging and for policies fast enough that the pause is negligible.

\paragraph{Asynchronous prefetch with linear-overlap blending.}
In asynchronous execution, inference and control run in parallel; \name implements the latency-aware overlap-blending scheme of~\citet{sima2026kai0}. A background thread keeps requesting new action chunks, while the main control loop executes actions from a buffer at the robot control rate. This avoids blocking the robot while the policy performs a forward pass, but introduces a temporal alignment problem. When a newly predicted chunk arrives, the robot may already have advanced several control steps since the observation that produced it. The early actions in the chunk are therefore stale, and the remaining actions may not align smoothly with the buffered trajectory already being executed.

\name addresses this issue with a latency-aware linear-overlap buffer. Let $k$ denote the number of control steps executed since the previous chunk was integrated, which estimates the inference delay, and let $k_{\max}$ be the maximum number of old actions to discard. When a new chunk $A^{\text{new}}$ arrives, the buffer first removes its leading $\min(k, k_{\max})$ actions, since these actions correspond to timesteps that the robot has likely already passed. After this trim, $a^{\text{old}}_i$ and $a^{\text{new}}_i$ index the same absolute control timestep. The remaining actions are then merged with the current buffer $A^{\text{old}}$ over an overlap window of length $L = \min(|A^{\text{old}}|, |A^{\text{new}}|)$, so that $L$ never exceeds the prediction horizon $H$:
\begin{equation}
a^{\text{blend}}_i = w_i\, a^{\text{old}}_i + (1 - w_i)\, a^{\text{new}}_i,\quad
w_i = 1 - \frac{i}{L-1},
\end{equation}
where $i = 0, \dots, L-1$. This weighting starts from the buffered trajectory and gradually shifts to the new prediction, reducing discontinuities at chunk boundaries. Actions of $A^{\text{new}}$ beyond the overlap window are appended unchanged. Figure~\ref{fig:inference-strategy}(a) illustrates this process: the leading \emph{disposed} actions are trimmed, the overlap window is \emph{fused} by the linear blend, and the \emph{fresh tail} extends the trajectory.

\paragraph{Temporal ensembling (ACT-style).}
Following \citet{act}, temporal ensembling smooths asynchronous action chunks by aggregating predictions in absolute time. Instead of replacing previous predictions, each action is assigned to its target control timestep, and all predictions for the same timestep are retained. Let $\mathcal{P}(t)=\{(j,a)\}$ denote the predictions associated with timestep $t$, where $j$ indexes the inference call and smaller $j$ indicates an earlier call. The executed action is computed as
\begin{equation}
\bar{a}_t =
\frac{\sum_{(j,a)\in\mathcal{P}(t)} e^{-m\rho_t(j)} a}
{\sum_{(j,a)\in\mathcal{P}(t)} e^{-m\rho_t(j)}} ,
\end{equation}
where $\rho_t(j)$ is the local position of call $j$ within $\mathcal{P}(t)$, counted from $0$ for the oldest call and increasing toward the newest. Thus, earlier predictions receive larger weights, while later predictions are exponentially discounted. This up-weighting of older predictions is inherited from ACT's temporal-ensembling default rather than a design choice specific to \name. The coefficient $m$ controls the decay strength, with larger values emphasizing the earliest prediction and the default $m=0.01$ producing nearly uniform averaging. Predictions beyond a fixed temporal horizon are pruned for efficiency. Since each chunk may cover multiple future timesteps, the background worker integrates the full, uncropped chunk to maximize temporal overlap. Figure~\ref{fig:inference-strategy}(b) shows the resulting overlap structure, where every chunk active at an \emph{ensemble slice} contributes to the executed action.

\paragraph{Naive asynchronous replacement.}
A minimal baseline: the most recent chunk simply replaces the buffer, indexed by a global timestep so that the action executed corresponds to the elapsed time since the chunk was produced, which compensates for latency by indexing rather than by blending. It is useful as a control to isolate the contribution of smoothing.

\paragraph{Real-Time Chunking (RTC).}
Real-Time Chunking~\citep{rtc} reduces seams at the source rather than post-hoc. Each inference request carries the previously committed action chunk back to the server, shifted forward by the latency $k$ and with the trailing entries padded with the last committed action. Generation is thereby conditioned to remain consistent with what the robot is already executing. The division of labor is deliberate: the client only performs the latency shift and attaches the prior actions to the request, while the server owns the delta conversion, normalization, and conditioned generation. The returned chunks are still passed through the linear-overlap buffer for a final smoothing pass. This double smoothing is inexpensive: when the server's conditioning succeeds, successive chunks already agree over the overlap window, so the linear blend is near-identity and acts only on any residual boundary disagreement. Conceptually, RTC moves part of the smoothing burden into the model, while the client retains a robust fallback.

\paragraph{Debuggability.} Each of these strategies is fully logged through the three action streams of Section~\ref{sec:eval}, so the effect of a given smoothing method, including any latency artifacts it introduces, can be inspected after the run rather than guessed at on the robot.

\section{Data Collection}
\label{sec:collect}

The operation model of Section~\ref{sec:modes} is oriented around driving a trained policy, but the same console, robot descriptions, and transport backends also support the inverse problem of producing the demonstrations a policy is trained on. \name's Collect mode records teleoperated episodes directly into a training-ready dataset, using the same console, robot descriptions, and transport backends as deployment. Unlike the other modes, Collect runs no policy at all: a human drives the robot and the client only records.

\paragraph{Teleoperated recording.} An operator drives the robot by teleoperation, moving a leader arm that the follower mirrors, and \name records each frame as a synchronized pair: the robot's measured state and the commanded action, each available as joint angles and, via forward kinematics, end-effector pose. Recording is gated behind an explicit activation step, so capture never begins accidentally. Because both the state and the action are stored, a recording is simultaneously replayable for review and usable as a supervised training target. Live capture is implemented on the ROS1, ROS2, and ZMQ transports; a configuration becomes a collection configuration by declaring a recording schema, which is what activates the Collect tab.

\paragraph{Training-ready output.} Episodes are written in the LeRobot format~\citep{lerobot}, the same on-disk layout \name reads back for offline replay (Section~\ref{sec:backends}), with one dataset per task: per-step observations and actions in a columnar table, one H.264 video per camera, and dataset metadata. To satisfy the format's fixed-rate assumption, per-frame timestamps are synthesized at exact intervals while the jittery real capture time is preserved in a separate field. Saving is handed to a background writer so that live capture never stalls, and recordings are appended rather than overwritten.

\paragraph{Quality control.} Every finished episode is checked frame by frame for non-monotonic timestamps, missing or malformed camera frames, wrong-dimension vectors, non-finite values, and video/table length mismatches, then flagged as clean or problematic. Flagged episodes are never silently discarded: offending fields are zero-filled to keep the table regular and the issue is recorded for review, so bad data is quarantined rather than lost. The operator then reviews an episode by replaying it open-loop in the console, with no robot or policy server required, and records a pass/fail verdict with an optional free-text note, written in place without touching the recorded trajectory.

\section{Evaluation and Logging}
\label{sec:eval}

If deployment is the least consolidated part of the VLA stack, evaluation is the least systematic. Teams routinely judge a new checkpoint by running it a few times and forming an impression, a process that is subjective, unrecorded, and impossible to reproduce or audit. EVA-Client treats evaluation and logging as a core subsystem rather than an afterthought, so that ``is this checkpoint better?'' becomes a recorded, comparable measurement.

\paragraph{Why this is hard on real robots.} Unlike simulation, a physical evaluation cannot be replayed. The scene drifts between trials, a single overall impression is a subjective way to score a run, and the most informative signal, what the policy actually saw and did, is lost once the run ends unless it is captured. A credible evaluation harness must therefore standardize the trial protocol, score each trial against explicit milestones rather than a single overall judgment, and persist enough information to reconstruct and compare runs after the fact.

\paragraph{Scored trials and scenes.} Evaluation is organized into scenes, each pairing an object configuration with a prompt and a per-trial list of milestone outcomes. Each scene is run for a configured number of trials, and every trial is scored against its milestones, yielding a graded process score rather than just binary success, with an optional free-text note. Every trial is persisted as a structured record, one per trial, keyed by (scene, position, trial) and carrying the milestone outcomes, graded score, duration, and free-text note. Each record is bound to a recorded video through a clip identifier minted when the run starts, so the video and its score stay linked even across restarts. From these records, per-scene and per-checkpoint statistics such as success rate and process score are computed from the logs rather than reduced to a single anecdotal outcome.

\paragraph{Multi-checkpoint comparison.} Several checkpoints can be evaluated within one session. Switching the ``active model'' transparently re-targets the policy endpoint and re-runs warmup, so each checkpoint is scored on the same scenes and milestones under identical conditions. Results are logged per checkpoint and kept separate, so they can be lined up side by side in the result viewer afterward.

\paragraph{Complete Logging of the Three Action Streams.} The substrate beneath evaluation is exhaustive logging. For every run, EVA-Client records three parallel, timestamped action streams covering the raw per-chunk predictions returned by the policy, the smoothed actions after the inference strategy's buffering, and the executed actions actually sent to the robot. Each step is tagged with its source chunk. Recording all three, rather than only the final command, makes the inference pipeline debuggable. Plotting them together reveals chunk boundaries, overlap regions, and latency compensation directly, and it makes it possible to attribute a bad motion to the model, the smoothing, or the execution layer. On the executed stream, the narrow transition zone where two chunks overlap is visible directly.

\paragraph{Read-only result viewer.} Logged results are explorable through a read-only web viewer that connects to neither robot nor policy server. It renders per-checkpoint statistics and per-scene breakdowns, and provides a side-by-side comparison mode across checkpoints, with each scored trial linking to its recorded video and milestone outcomes. The three action streams are exported as per-run tables organized by prompt, mode, and strategy, with chunk ownership tracked per step so overlap regions can be visualized. Because the viewer consumes only the persisted logs, evaluation results can be inspected, shared, and re-analyzed long after the robot is powered down. This closes the loop from a physical run to an auditable, reproducible record.

\section{Limitations and Roadmap}
\label{sec:roadmap}

\paragraph{Limitations.} EVA-Client is deployment infrastructure, not a policy or a benchmark: it does not train models, and the qualitative task outcomes reported here are illustrative observations from real deployments rather than a controlled study. Live transport currently covers ROS1, ROS2, and the ZMQ socket backend, but cameras on non-ROS robots still depend on a middleware-specific image source. The inverse-kinematics solver targets serial-arm manipulators; other morphologies are supported only through their own robot descriptions.

\paragraph{Roadmap.} Several directions build on the deployment substrate already in place.

\begin{enumerate}
    \item \textbf{Reinforcement-learning data collection.} We aim to close the loop from evaluation back to external training, treating a policy's rollouts and their reward or outcome labels as material for reinforcement learning and interactive fine-tuning rather than passive logs. Human-in-the-loop collection fits here: an operator takes over or nudges the arm mid-rollout, and the correction is recorded as targeted training signal.

    \item \textbf{Agentic policies.} Beyond a single flat policy, we aim to make \name the runtime for hierarchical policies, where a high-level planner or vision-language agent dispatches sub-goals to low-level controllers. Since the interfaces are already narrow, the same client can host the controllers while exposing hooks for a planner to issue sub-goals and read back execution state.

    \item \textbf{Data annotation.} We plan to extend Collect mode with fine-grained task and sub-task annotation, segmenting long-horizon episodes into labeled sub-task units and milestones within the same LeRobot dataset. This makes a collection reusable at the level of individual manipulation phases.

    \item \textbf{Broader robot morphologies.} Answering the serial-arm limitation above, we aim to widen the supported robot set to humanoid and mobile-base morphologies by adding the robot descriptions and kinematic models each requires, registered through the same interface as the existing arms.
\end{enumerate}

Across these directions, as with our current collection modes, we release the tooling that produces and labels the data, not the data itself.

\section{Conclusion}
\label{sec:conclusion}
This report positions \name{} as the deployment counterpart to an otherwise increasingly mature policy-training ecosystem. By decoupling signal sources, robot backends, action spaces, and inference strategies within a unified client, \name{} turns real-robot deployment from a collection of one-off scripts into an inspectable, reproducible, and comparable process. The same substrate supports teleoperated data collection, scored evaluation, exhaustive action logging, and post-hoc result replay, thereby linking data collection, policy execution, and physical evaluation into a single closed loop that feeds back into external training. EVA-Client is not a policy, a benchmark, or a dataset; it is an open deployment, collection, and evaluation layer designed to give diverse models and robot platforms a shared foundation for real-robot experimentation.

\section*{Author List}
\label{sec:author}

\begingroup
\makeatletter
\long\def\@makefntext#1{%
  \parindent 0pt%
  \noindent
  \hangindent=1.8em%
  \hangafter=1%
  \makebox[1.8em][c]{\@makefnmark}#1%
}
\makeatother

\renewcommand{\thefootnote}{%
  \ifcase\value{footnote}%
  \or *%
  \or \textdagger%
  \or \Letter%
  \else \arabic{footnote}%
  \fi
}
\setcounter{footnote}{0}

\subsection*{Contributors}
Heqing Yang\footnote{Core Contributors \& Equal Contribution.}, Yang Yi\footnotemark[1], Liyao Wang, Linqing Zhong, Donglin Yang, Ruipu Wu, Zitong Bai

\textbf{Framework construction:} Heqing Yang\footnotemark[1], Yang Yi\footnotemark[1]

\textbf{Robot hardware adaptation:} Heqing Yang (Agilex, Franka, Arx), Yang Yi (Arx, Agilex), Liyao Wang (Galaxea, UR5e), Linqing Zhong (AgiBot G2), Donglin Yang (Franka), Ruipu Wu (AgiBot G2)

\textbf{Writing and illustration:} Yang Yi, Heqing Yang

\textbf{Website and document:} Heqing Yang, Zitong Bai, Yang Yi

\subsection*{Team Leads}
Fengjiao Chen\footnote{Project Lead.}, Manyuan Zhang, Linjiang Huang\footnote{Corresponding Authors: \texttt{\{ljhuang, liusi\}@buaa.edu.cn}}, Si Liu\footnotemark[3]

\endgroup

{\small
\bibliographystyle{\colabbibstyle}
\bibliography{eva_report}
}

\end{document}